\title{On Automation and Medical Image Interpretation, With Applications for Laryngeal Imaging}
\author{
        Habib J. Moukalled \\
        Computer Science and Engineering Department \\
        University of South Carolina\\
        Columbia, SC 29208, U.S.A. \\
	\url{habib.moukalled@gmail.com}
}

\date{\today}

\documentclass[11pt]{article}

\usepackage{graphicx}

\usepackage{tikz}
\usetikzlibrary{arrows,automata}
\usetikzlibrary{external}
\usetikzlibrary{positioning}

\usepackage{amsfonts}
\usepackage{amsthm}
\usepackage{amssymb}
\usepackage[cmex10]{amsmath}

\PassOptionsToPackage{hyphens}{url}\usepackage{hyperref}
\usepackage{breakurl}

\usepackage[margin=1.0in]{geometry}

\begin{document}
\maketitle

\begin{abstract}
Indeed, these are exciting times. We are in the heart of a digital renaissance. Automation and computer technology allow engineers and scientists to fabricate processes that amalgamate quality of life. We anticipate much growth in medical image interpretation and understanding, due to the influx of computer technologies. This work should serve as a guide to introduce the reader to core themes in theoretical computer science, as well as imaging applications for understanding vocal-fold vibrations. In this work, we motivate the use of automation and review some mathematical models of computation. We present a proof of a classical problem in image analysis that cannot be automated by means of algorithms. Furthermore, discuss some applications for processing medical images of the vocal folds, and discuss some of the exhilarating directions the art of automation will take vocal-fold image interpretation and quite possibly other areas of biomedical image analysis.
\end{abstract}

\smallskip
\textbf{Keywords} \small Computability, laryngeal imaging, medical image interpretation, pattern simulation, snakes. \\

\normalsize

\medskip
\noindent \textbf{Subject Classification:} Medical Image Interpretation.

\section{Introduction}
Although more recently it has been recognized as its own discipline, the fruits of Biomedical Engineering (BME) have long been apparent, and will continue to revolutionize the way human beings understand and administer medicine. As a multi-disciplinary field, BME draws upon the expertise of Biology, Chemistry, Engineering, Mathematics, Medicine, Physics, Physiology, Probability and Statistics, and several others. Advancements in computer technologies and imaging modalities over the last several decades have attributed to the wide spread growth of medical image analysis, an application central to BME \cite{nebeker:02}. Semiautomated algorithms will pave the way to better understanding of medical image analysis and help shape automated biomedical imaging.

In the era of big data and information processing, we are in the heart of a digital renaissance. Despite the obvious power of the computer technology available today, there exist many inherent limitations that dramatically affect the types of software and methodology available for medical image analysis. Furthermore, several researchers have posed general questions regarding the efficacy of automated systems. For example, trust, workload, and risk influence the adoption of automated systems, while individual differences make predicting success of these systems difficult, and false alarms lead to underutilization of the automated systems \cite{parasuraman:97}.

In medical image analysis, automation should serve as a tool to aid clinicians, pathologists, physicians, and surgeons, not serve as a replacement. Deployment of robust automated systems takes several years of research and development, and several validation studies to ensure the quality of the automated work. For example, the air traffic control, railroad systems, and several areas of manufacturing have been successful in harvesting the fruits of automation. However, in medical image analysis the fruits of automation are carried by quite a different tree, due to the very nature of each specialist's function along the medicinal pipeline. 

Researchers have proposed breaking down automated functions into four classes ($1$) information acquisition, ($2$) information analysis, ($3$) decision and action selection, and ($4$) action implementation \cite{parasuraman:00}. From our experience, categories ($2$) information analysis and ($3$) decision and action selection, for the time being, are difficult to achieve without human interaction in medical image analysis, and we expect to see them evolve and grow in the coming years. Category ($2$) requires cognitive capabilities such as working memory and inferential processes, while ($3$) requires decisions yielding different outcomes while having insight on the consequences of such actions \cite{parasuraman:00}. 

Engineers seeking to automate systems must proceed carefully as automation has been shown to alter human behavior, since automation can make it difficult for the automator and/or supervisor to predict or anticipate outcomes of scenarios \cite{parasuraman:97, parasuraman:00, parasuraman:08}. Furthermore, an additional factor to consider when employing automated systems is the effect the automation will have on human decision making \cite{skitka:99}. Automated agents are already becoming viable teammates to humans in the workforce. Although humans naturally behave socially to their digital counterparts, human-machine trust deteriorates faster than human-human trust, but can be greatly reduced when human users and supervisors better understand the workings of their automated teammates \cite{madhavan:04}.


\section{The Art of Automation}

In this section, we provide the reader with a brief overview of two well-studied abstract machines, models of computation. Finite automata (limited memory computers) and Turing machines, which are more representative of the computers we use in every day life. To make the illustration of the concepts in sections $2$ and $3$ more tangible, we have developed a software package containing simulated examples allowing the reader to follow along at home. See: \url{https://github.com/habisoft/} for a listing of our software repository corresponding to this paper.

\subsection{Automata Theory at a Glance}

\textit{Finite Automata} are mathematical models of simple computation. Despite their limited amounts of memory, they play a key role in several practical areas of computer science, such as, compiler design, command line and interpreter design, programming language development, hardware design, and other various text processing applications \cite{sipser:06,lewis:98}. Finite automata have long been studied for their ability to generate and recognize \textit{regular languages}. Regular languages have several special properties, such as, their closure properties, that is, the resulting language is also a regular language when performing boolean operations like complementation, intersection, and  union on the strings of the language in question. Furthermore, when performing regular operations like concatenation, Kleene star, and string reversal, the resulting string will still be members of a regular language. 

Mechanically speaking, finite automata are modeled as a reading head and a finite length input tape. The reading head of a finite automaton contains a finite number of states for defining its control logic, and can only move from left to right while scanning across its input tape, reading one input symbol at a time. Depending on the sequence of input symbols read, and the finite automaton's transition function, the automaton will transition to different states. A finite automaton is not allowed to write to its input tape, this can be viewed as read-only memory (ROM). On any given input string, a finite automaton will \textit{accept} the string if the string is a member of the language that the automaton recognizes or \textit{reject} the string if the string is not a member of the language the automaton recognizes. Fig. $1$ illustrates a mechanical diagram of a finite automaton. 

\begin{figure}[!ht]
\centering
\includegraphics[width=6.0cm]{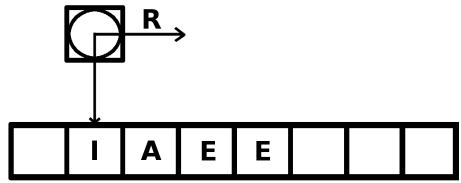}
\caption{Mechanical diagram of a finite automaton. Reading one element at a time from a finite length tape, the reading head moves from left to right.}
\label{fig:Fig1.1}
\end{figure}

Suppose we have a finite alphabet of input symbols $\Sigma = \{\textnormal{a}, \textnormal{b}, \textnormal{c}, \hdots, \textnormal{z}, \textnormal{A}, \textnormal{B}, \textnormal{C}, \hdots, \textnormal{Z}\}$, that is, the alphabet of lower case letters a through z and upper case letters A through Z from the English alphabet. Note, by definition, all strings from $\Sigma$ are of finite length \cite{sipser:06, lewis:98}. Motivated by Cornell University's library of e-print articles at \url{arxiv.org} in the Computing Research Repository (CoRR), we devise the following \textit{regular expression} (REX) that generates the language of all strings that contain at least one occurrence of the substring CoRR is given by Eq. ($1$)

\begin{equation}
C_{\textnormal{REX}} = \Sigma^*(\textnormal{CoRR})^+\Sigma^*.
\end{equation}

\noindent In Eq. ($1$), the $\Sigma^*$, is an application of the \textit{Kleene star} operation over the alphabet $\Sigma$, and denotes $0$ to $\infty$ many strings from the alphabet $\Sigma$. The term $(\textnormal{CoRR})^+$ represents $1$ to $\infty$ repetitions of the string $\textnormal{CoRR}$, and is an example using the \textit{Kleene plus} operation over the string CoRR. Now, in plain English, the regular expression in Eq. ($1$) can be read as any string (including the empty string), followed by the string CoRR (at least once), followed by any string (including the empty string). Let this language be called substring CoRR.

Fig. $2$ is a \textit{deterministic finite automaton} (DFA) $D$ that recognizes the language substring CoRR, which we denote as $L(D) = C_{\textnormal{REX}}$, that is, the \textit{language} of $D$ is $C_{\textnormal{REX}}$. 

\begin{figure}[!ht]
\begin{center}
\begin{tikzpicture}[>=stealth',shorten >=1pt,auto,node distance=2.5cm]
  \node[initial,state,minimum size=.05cm, font=\footnotesize]		(q0)      {$q_0$};
  \node[state,minimum size=.05cm, node distance=2.5cm, font=\footnotesize]	(q1) [right of=q0]  {$q_1$};
  \node[state,minimum size=.05cm, font=\footnotesize]			(q2) [right of=q1] {$q_2$};
  \node[state,minimum size=.05cm, font=\footnotesize]			(q3) [right of=q2] {$q_3$};
  \node[accepting,state,minimum size=.05cm, font=\footnotesize]	(q4) [right of=q3] {$q_4$};

  \path[->] (q0)  edge [loop above, distance=35, looseness=3, font=\footnotesize] node {$\Sigma \setminus \{\textnormal{C}\}$} (q0)
             edge [thick]            node {\textbf{C}} (q1)
	(q1)  edge [loop above, distance=30, looseness=3, thick] node {\textbf{C}} (q1)
        (q1) edge [above, bend right, distance=40, looseness=2, font=\tiny]  node {$(\Sigma \setminus \{\textnormal{C}\}) \setminus \{\textnormal{o}\}$} (q0)
             edge [thick]      node {\textbf{o}} (q2)
	     
        (q2)    edge [below, bend left, distance=25, font=\footnotesize]  node {$\Sigma \setminus \{\textnormal{R}\}$} (q0)
	(q2) edge [thick]	node {\textbf{R}} (q3)

	(q3)    edge [below, bend left, distance=100, looseness=1, font=\footnotesize]  node {$\Sigma \setminus \{\textnormal{R}\}$} (q0)
	(q3)	edge [thick]	node {\textbf{R}} (q4)

	(q4)  edge [loop above, font=\footnotesize] node {$\Sigma^*$} (q4);
\end{tikzpicture}
\end{center}
\caption{A \textit{Deterministic finite automaton} $D$, such that $L(D) = C_{\textnormal{REX}}$, that is, $D$ recognizes the language substring CoRR, with accepting paths in bold.}
\label{fig:Fig1}
\end{figure}
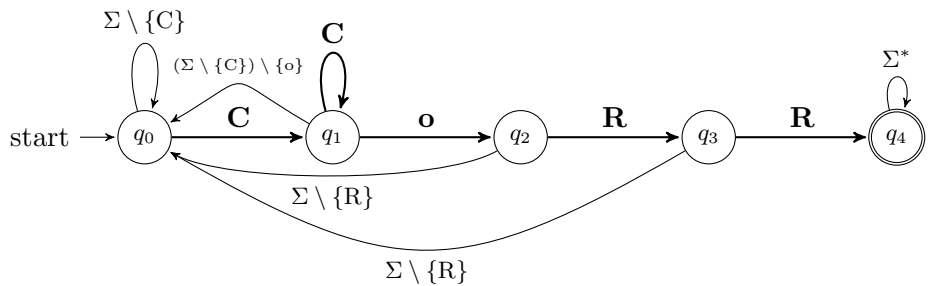

\noindent We illustrate DFAs as directed graphs. When DFA $D$ is in state $q_0$ and it reads a C, the DFA will transition into state $q_1$, when in state $q_1$, if an o is read, the DFA will transition into state $q_2$, so on and so forth. Upon reading C and o, and a total of two Rs are read, the DFA will be will in state $q_4$, the accepting state. The accepting state is denoted as a circle within a circle. The attribute that makes an automaton deterministic is the fact that a single transition (edge) is given for each input symbol of the alphabet $\Sigma$ for each state of the automaton. In Fig. $2$, for simplicity, we do not draw a transition for each lower case and capital letter of the English alphabet, rather we write $\Sigma \setminus \{\textnormal{a particular symbol}\}$. For example, state $q_0$ has a reflexive edge that points right back into state $q_0$ with the transition labeled $\Sigma \setminus{\{\textnormal{C}\}}$, which is read as the entire alphabet \textit{except} for C. Likewise, the edge from $q_1$ to $q_0$ is read as any symbol in the alphabet $\Sigma$, except for C and o.

\subsection{Computability Theory at a Glance}

Similar to a finite automaton, a \textit{Turing Machine} (TM), first introduced by Alan Turing in the late $1930$'s as the \textit{a-machine}, is a model of computation \cite{turing:36}. Like the automaton in Fig. $1$, a TM has a reading head and an input tape. Unlike finite automata, the TM's reading head is a read/write head, and the input tape is of infinite length. Furthermore, the TM's reading head can move left or right as determined by the device's transition function. Unlike finite automata, a TM has a separate start and accept states; they take place immediately. These subtle differences between finite automata and TMs creates a world of difference in terms of the types of languages these devices can recognize. 

On any input string, a Turing machine will \textit{accept}, \textit{reject}, or \textit{loop}. Accepting and rejecting are halting behaviors, the TM will start its simulation and stop after a finite number of steps. A language is \textit{Turing-decidable} or \textit{recursive} if some Turing machine halts and accepts every string in the set or halts and rejects for non members. A language is \textit{Turing-recognizable} or \textit{recursively-enumerable} if some TM enumerates the strings in a language, but does not necessarily halt. We note, the computers we employ in everyday life do not have infinite amounts of memory, however, the amount of memory available is so large, that modeling a computer as finite automata is inaccurate and counterproductive \cite{hopcroft:01}. In fact, our intuitive notion of algorithms for the computers we use everyday are equivalent to Turing machine programs \cite{sipser:06}.

We use set notation to denote sets of strings that obey some property, we call these sets \textit{languages}. The following is the set theoretic notation for the language $C_{\textnormal{REX}}$, the language of strings containing at least one occurrence of CoRR.

\begin{center}
\indent $C_{\textnormal{REX}} =$ \hspace{2pt} \{$w \hspace{2pt} | \hspace{1pt} w$  is a string that contains the substring CoRR \}. \\
\end{center}

\noindent Below, the TM $U$ is a decider for the language $C_{\textnormal{REX}}$. $U$ is an example of the \textit{universal} Turing machine and is capable of simulating finite automatons and other Turing machines. In this particular example, $U$ will simulate the DFA $D$ on the input string $w$, in this case, $D$ can be thought of as a subroutine or function in regards to modern programming. \\

\noindent \textnormal{$U =$ `` on input $<D, w>$, where $D$ is a string encoding the \\
\indent description DFA $D$ and $w$ is a string: \\
\indent ($1$.) Simulate DFA $D$ on the input string $w$. \\
\indent ($2$.) If $D$'s simulation ends in an accepting state, \textbf{accept}, \\
\indent \indent if the simulation ends in a non accepting state, \textbf{reject}. "} \\

\subsection{On Computability and Countability}

The very framework of computability and to a large degree most of discrete mathematics are built upon set theory and countability of sets. One interesting result of computability is that the set of all strings $\Sigma^*$ can be put into \textit{correspondence} with the set of natural numbers $\mathbb{N}$, therefore, $\exists$ a \textit{bijection} $z:\Sigma^* \rightarrow \mathbb{N}$, a function that maps every element of $\Sigma^*$ to every element of $\mathbb{N}$. For example, suppose our finite alphabet is the binary alphabet $\Sigma = \{0, 1\}$. Our bijection $z: \Sigma^*\rightarrow \mathbb{N}$ goes as follows, in \textit{lexicographical} order list all the strings in $\Sigma^*$, that is, in increasing order, list all strings of length $0$, then all strings of length $1$, then all strings of length $2$, so on and so forth. While listing all strings in lexicographical order list all the numbers in $\mathbb{N}$. As the elements of both sets are being enumerated, pair the current element of $\Sigma^*$ with the current element in $\mathbb{N}$. Below, a visual representation of the bijection $z:\Sigma^* \rightarrow \mathbb{N}$ is given, where $\epsilon$ is the empty string.

\abovedisplayskip=0pt
\begin{align*}
\Sigma^* &= \{\epsilon,& &0,& &1,& &00,& &01,& &10,& &11,& &000,& &\hdots\} \\
\mathbb{N} &= \{0,& &1,& &2,& &3,& &4,& &5,& &6,& &7,& &\hdots\}
\end{align*}

\noindent Via the sets enumerated above, each column in $\Sigma^*$ is aligned to a corresponding column in $\mathbb{N}$, this is our bijection, clearly the set of all strings and the set of natural numbers are both countably infinite, therefore, the sets are the same size or \textit{cardinality}, denoted $\left|\Sigma^*\right| = \left|\mathbb{N}\right|$. In fact, bijective functions are \textit{computable maps}, functions that are decidable by some TM in a polynomial number of steps with respect to the length of the given input. Georg Cantor, arguably the father of elementary set theory, spent a major portion of his career working on the cardinality of various infinite sets and published many works on the topic in the late $1800$'s. One interesting problem Cantor studied was the cardinality of the rational fractions $\mathbb{Q} = \{\frac{m}{n} \hspace{1pt} | \hspace{1pt} m, n \in \mathbb{N}\}$. Oddly enough, $\left|\mathbb{Q}\right| = \left|\mathbb{N}\right|$.

\newtheorem{theorem}{Theorem}[section]
\newtheorem{lemma2.1}[theorem]{Lemma}

\begin{lemma2.1}
\textnormal{$\exists$ a bijection $y:\mathbb{Q} \rightarrow \mathbb{N} \hspace{4pt} \therefore \hspace{4pt}  \left|\mathbb{Q}\right| = \left|\mathbb{N}\right|$, today the bijection is known as Cantor's \textit{diagonalization} argument \cite{sipser:06, hopcroft:01}, we will not prove it here, however, in the proof of the next lemma the same technique is applied.}
\end{lemma2.1}

\newtheorem{lemma2.2}[theorem]{Lemma}

\begin{lemma2.2}
Proof: \textnormal{$\exists$ a bijection $f:\mathbb{N \times N}\rightarrow\mathbb{N}$} 

\textnormal{In order to demonstrate the bijection from the Cartesian product of the natural numbers with the natural numbers we must pair each element from $\mathbb{N} \times \mathbb{N}$ with a \textit{unique} element of $\mathbb{N}$. We begin by listing all possible pairs of $\mathbb{N \times N}$ in an infinite table, which is given below.}

\[
\begin{bmatrix} 
(0, 0)_{0\phantom{0}} & (0, 1)_{2\phantom{0}} & (0, 2)_{5\phantom{0}} & (0, 3)_{9\phantom{0}} & (0, 4)_{14} & \hdots\phantom{_{00}} \\
(1, 0)_{1\phantom{0}} & (1, 1)_{4\phantom{0}} & (1, 2)_{8\phantom{0}} & (1, 3)_{13} & (1, 4)\phantom{_{00}} & \hdots\phantom{_{00}} \\
(2, 0)_{3\phantom{0}} & (2, 1)_{7\phantom{0}} & (2, 2)_{12} & (2, 3)\phantom{_{00}} & (2, 4)\phantom{_{00}} & \hdots\phantom{_{00}} \\
(3, 0)_{6\phantom{0}} & (3, 1)_{11} & (3, 2)\phantom{_{00}} & (3, 3)\phantom{_{00}} & (3, 4)\phantom{_{00}} & \hdots\phantom{_{00}} \\
(4, 0)_{10} & (4, 1)\phantom{_{00}} & (4, 2)\phantom{_{00}} & (4, 3)\phantom{_{00}} & (4, 4)\phantom{_{00}} & \hdots\phantom{_{00}} \\
\vdots\phantom{_{00}} & \vdots\phantom{_{00}} & \vdots\phantom{_{00}} & \vdots\phantom{_{00}} & \vdots\phantom{_{00}} & \ddots\phantom{_{00}}
\end{bmatrix}
\]

\noindent \textnormal{Notice in the table above, we use subscript notation to demonstrate the element of $\mathbb{N}$ that each ordered pair from $\mathbb{N \times N}$ maps to. The bijection works by starting at element $(0, 0)$ and then simply counting up along all the subsequent diagonals after the first diagonal starting at $(0, 0)$. More specifically, the bijection $f$, has the following closed formula:} 

\[
f(a, b) = \frac{(a + b)(a + b + 1)}{2} + b. \\
\]

\noindent \textnormal{The counting scheme presented in this lemma is a variation of the diagonalization argument, also known as Cantor's \textit{pairing function}, thus, we have successfully illustrated the existence of a bijection $f:\mathbb{N \times N}\rightarrow\mathbb{N} \hspace{4pt} \therefore \hspace{4pt} \left| \mathbb{N \times N}\right| = \left|\mathbb{N}\right|$} \qed
\end{lemma2.2}

Bijective functions have the special property of being \textit{invertible} functions. The concept of invertibility has deep implications, thus one problem instance (language) may be casted to another more malleable set in a finite number of steps. Another result of set theory is the notion of a set's \textit{complement}, which is a set of everything that is not contained in the set in question. For example, the complement of set $A$ is denoted as $\overline{A}$, and would contain every element not found in $A$. Another example would be the complement of $\mathbb{N}$, the real numbers, therefore, $\overline{\mathbb{N}} = \mathbb{R}$, these are decimal numbers that are not necessarily rational and in fact, $\mathbb{R}$ is an infinite set of infinite sets. Another interesting result where computability meets countability is that the set of all TMs is countably infinite, thus $\left|\textnormal{TM}\right| = \left|\mathbb{N}\right|$, all TMs can be put in correspondence with the set of natural numbers \cite{sipser:06, hopcroft:01}. Yet, this also has deep implications in that there are more languages (problems) than there are Turing machines, this in turn implies that there are some languages that are not even Turing-recognizable, which in turn brings us to the following lemma.

\newtheorem{lemma2.3}[theorem]{Lemma}

\begin{lemma2.3}
Proof: \textnormal{A language $L$ is decidable if and only if it is Turing-recognizable and its complement $\overline{L}$ is also Turing-recognizable (co-Turing-recognizable) \cite{sipser:06}.} \\

\noindent \textnormal{$\Rightarrow$ if $L$ is decidable, both $L$ and $\overline{L}$ are Turing-recognizable.} \\

\textnormal{Any decidable language $L$ is by default Turing-recognizable, and given the closure property of the set of decidable languages under complementation, the complement $\overline{L}$ is also decidable. Therefore, if $L$ is decidable, then the decidability and recognizability of $\overline{L}$ is trivially true.} \\

\noindent $\Leftarrow$ \textnormal{if both $L$ and $\overline{L}$ are Turing-recognizable.} \\

\textnormal{For this direction, we let the TM $M_1$ be the recognizer for $L$ and the TM $M_2$ be the recognizer for $\overline{L}$. Now, we construct the TM $M$ that decides $L$.} \\

\noindent \textnormal{$M =$ `` on input $w$, where $w$ is a string: \\
\indent ($1$.) Run both $M_1$ and $M_2$ on input $w$ parallel. \\
\indent ($2$.) If $M_1$ accepts, \textbf{accept}; If $M_2$ accepts, \textbf{reject}. "} \\

\noindent \textnormal{Running the TMs $M_1$ and $M_2$ in parallel can be achieved by simulating each machine's input tape on a two-tape TM that toggles between simulating a step on $M_1$ and then simulating a step on $M_2$ until either machines accept. All strings $w$ will be either in $L$ or $\overline{L}$. Therefore, $M_1$ or $M_2$ will accept $w$ at some point during their simulation. $M$ accepts all strings in $L$ and rejects all strings in $\overline{L}$. Therefore, $M$ will always halt, making it a decider for $L$.} \qed 

\end{lemma2.3}

Despite the obvious power of our computers, there are limitations on the types of problems algorithms can solve, and they should not be taken lightly. We live in a fast-paced society where time and money dominate the decisions we make in our waking lives, therefore, having a firm understanding of computationally infeasible problems can save time, money, and the stresses induced by exhausting either without some sort of gain.

\section{A Simple Energy Optimization Problem in Image Analysis}

In computer vision and image processing related tasks, \textit{segmentation} is a processing step that aims to separate or extract features of interest from image data. In many medical image analysis problems, segmentation is a preliminary step that must be performed before measurements and additional relationships can be determined. Some segmentation methods employ thresholding of intensity information, some take advantage of edge information, while some methods even use computational geometry to incorporate prior shape information. One popular tool in image processing is the \textit{active contour} or \textit{snake} model, which is capable of incorporating edge, intensity, and prior shape information to segment and extract features of interest from digital images by deforming a contour towards salient image features \cite{kass:88}. 

Since their introduction, several other various deformable models have been devised including level set and curve evolution algorithms that exploit intensity information by evolving a contour using the inherent topology information specified by the change of intensity within an image \cite{osher:88, bresson:07}. Some snake models even treat the contour of interest as foreground information contained within the domain of a closed-loop curve being segmented from the rest of the image, which is treated as background information, by considering the behavior of the intensity information contained within the closed curve \cite{chan:01}. While other methods choose to deform a finite element model that specifies the contour of interest \cite{cohen:90}. We focus on classical snake theory as its roots extend into modern renditions of these concepts.

\subsection{Snake-energy Optimization}

Our concern is discrete sets of pixels and the steps necessary to partition the data into meaningful regions. Let $I$ be a $2$D gray scale image with two dimensions corresponding to the width and height of the image, and one dimension corresponding to the intensity, such that $I \subseteq \mathbb{N} \times \mathbb{N} \times \mathbb{N}$, where the values of each dimension are bounded by the natural numbers. A snake $\mathbf{s}$ is a spline bounded by the image plane, therefore $\mathbf{s} \in I$. Over the years, the series of vertices that comprise the snake have been termed \textit{snaxels}, a short hand for snake elements. We denote the length of our snake as $|\mathbf{s}| = n$, where $\mathbf{s}$ has $n$ snaxels. In classical snake theory, a snake will satisfy the following force-balance condition $\mathbf{f}_{\textnormal{int}} + \mathbf{f}_{\textnormal{ext}} = 0$, where $\mathbf{f}_{\textnormal{int}}$ is an internal force, the result of the shape of the snake, and $\mathbf{f}_{\textnormal{ext}}$ is an external potential-force field derived from image data. Until the force-balance condition is met, our snake $\mathbf{s}$ is repeatedly deformed to minimize an energy functional of the following form 


\begin{equation}
E^*_{\textnormal{snake}}(\mathbf{s}) = \sum^n_{i = 1} \left[ E_{\textnormal{int}}(\mathbf{s}_i) + E_{\textnormal{ext}}(\mathbf{s}_i)\right],
\end{equation}

\noindent where the internal force provided by our snake's shape is, 

\begin{equation}
E_{\textnormal{int}}(\mathbf{s}_i) = \frac{1}{2} \left[\alpha \cdot \left|\mathbf{s}_i - \mathbf{s}_{i - 1} \right|^2 +
\beta \cdot \left|\mathbf{s}_{i + 1} - \mathbf{s}_i + \mathbf{s}_{i - 1} \right|^2 \right],
\end{equation}

\noindent and the external potential provided by our image-data term is

\begin{equation}
E_{\textnormal{ext}}(\mathbf{s}_i) = -\gamma \cdot \left| \nabla I(\mathbf{s}_i) \right|^2.
\end{equation}

\noindent Our snake $\mathbf{s}$ is a controlled-continuity spline, that behaves like a rigid rubber band structure. In Eq. ($3$), $\alpha$ and $\beta$ are parameters that influence the continuity and rigidity, respectively. And Eq ($4$) is simply the magnitude of the image gradient where $\gamma$ is a weighting parameter for adjusting the strength or importance of the data term. Furthermore, several authors have improved the variational snake-deformation algorithm by devising external force fields that improve snake convergence even when the snake is far from the features of interest, by modeling the external potential as specific kinds of physical processes in time \cite{xu:98, bing:07, zhu:08}.

Variational approaches make use of higher order derivatives in order to compute internal and external forces acting upon a snake \cite{kass:88}. Yet, our data is discrete; in variational approaches, all is fine during snake-deformation until we wish to employ \textit{hard-constraints} on our contour of interest. Hard-constraints in optimization problems are those which force a solution, in our case, our snake, to obey some desired behavior. Under hard-constraints, variational approaches become numerically unstable. For example, a hard-constraint could be the specification of a desired distance between the snaxels that comprise a snake \cite{amini:90}. Furthermore, upon each successful iteration of snake-deformation, the external force acting upon a snake will exhibit minor perturbations as decimals values are influenced by noise and fluctuations in the computer memory \cite{amini:90, williams:92}. 

We note an example of a variational snake-deformation algorithm that should be able to overcome the pitfalls of classical snake-deformation approach, namely the numerical instability under hard constraints \cite{lobregt:95}. The method treats the series of snaxel's that comprise a snake as a set of masses connected by spline segments, that are accelerated towards image features. For each snaxel, an appropriate mass and dampening force can be applied, however, in our experience, selection of parameters that lead to ideal convergence can be difficult and the snake deformation can be time consuming. We wish to employ the discrete dynamic programming algorithm outlined in \cite{amini:90} as the kernel of the algorithms in our discussion on snake-energy optimization. 

Dynamic programming shares an interesting relationship with variational problems. It works by decomposing the system into a series of overlapping subproblems and enables us to optimize the cost of the variational system over a single dimension by treating a variational problem as a discrete multi-stage decision making process. When applying dynamic programming, convergence is guaranteed, since the number of possible states is a monotonically decreasing function with respect to an increase in iterations. Dynamic programming is capable of bypassing local minima. With regards to snake-energy minimization, such algorithms have these additional advantages: numerical stability due to not employing higher order derivatives in assessing the cost function, and guarantees the globally optimal snake-deformation and energy minimization for the given set of parameters. For more details, refer to \cite{amini:90, bellman:03}. Furthermore, in our experience, the dynamic programming approach is less sensitive to choice snake parameters in comparison to its variational counterparts.

\subsection{Application of Snakes in Laryngeal Imaging}

Here we provide the reader with some examples of snakes being used for medical image feature extraction, namely in regards to \textit{laryngeal high-speed videoendoscopy}, high-speed videos of vocal-fold vibrations \cite{deliyski:08}. \textit{Kymography}, with respect to vocal-fold (\textit{laryngeal}) imaging, is the use of high-speed line-scan images across the left and right vocal fold \cite{svec:96}. Typically, segmentation of the \textit{glottis}, the anatomical structure formed between the vocal folds as air separates them, is performed to determine additional characteristics such as measuring the left and right vocal folds. Several methods exist in image processing for segmenting features of interest from images, however, methods relying purely on image intensity information can result in imperfections and discontinuities in the extracted features. In biomedical imaging, structures of interest are subtle and can vary significantly from subject to subject, and it is for this reason many researchers choose to employ snakes or similar smooth contour delineation techniques, since this type of processing imposes a degree of smoothness on the extracted features.

Laryngeal imaging differs significantly from other types of medical images since the phenomenon of interest is non-stationary and is rapidly changing.  Vocal fold vibrations are fast, complex and fine. Several researchers \cite{marendic:01, allin:04, lohscheller:04, manfredi:06, karakozoglou:2012}, plus many others, have used snakes for laryngeal image analysis and its applications. Our goal here is not to provide a quantitative evaluation of such techniques, rather it is to introduce the reader to application of such image processing tools for medical image interpretation with respect to such laryngeal images. Fig. $3$ shows a pair of temporal snakes being deformed within the space-time domain of kymographic image sequences derived from laryngeal high speed video data \cite{moukalled:09}. 

\begin{figure}[!ht]
\centering
\includegraphics[width=8.0cm]{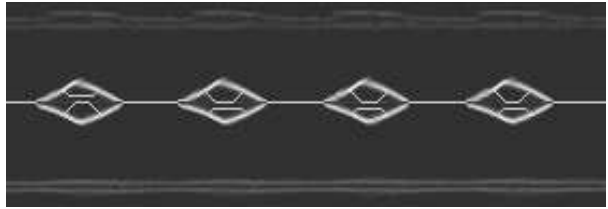}
\caption{Starting from the middle of the glottis along all glottal regions within the kymographic image, a pair of temporal snakes being deformed on a kymographic image.}
\label{fig:Fig3.1}
\end{figure}

\noindent Once the temporal-snake transform has been applied to all of the kymographic images containing the vocal folds, the resulting contours can be trivially remapped to the spatial domain of the recording as delineated left and right vocal folds shown in Fig. $4$. Note, in healthy vocal fold vibration, the vocal folds are restricted to moving left and right.

\begin{figure}[!ht]
\centering
\includegraphics[width=16cm]{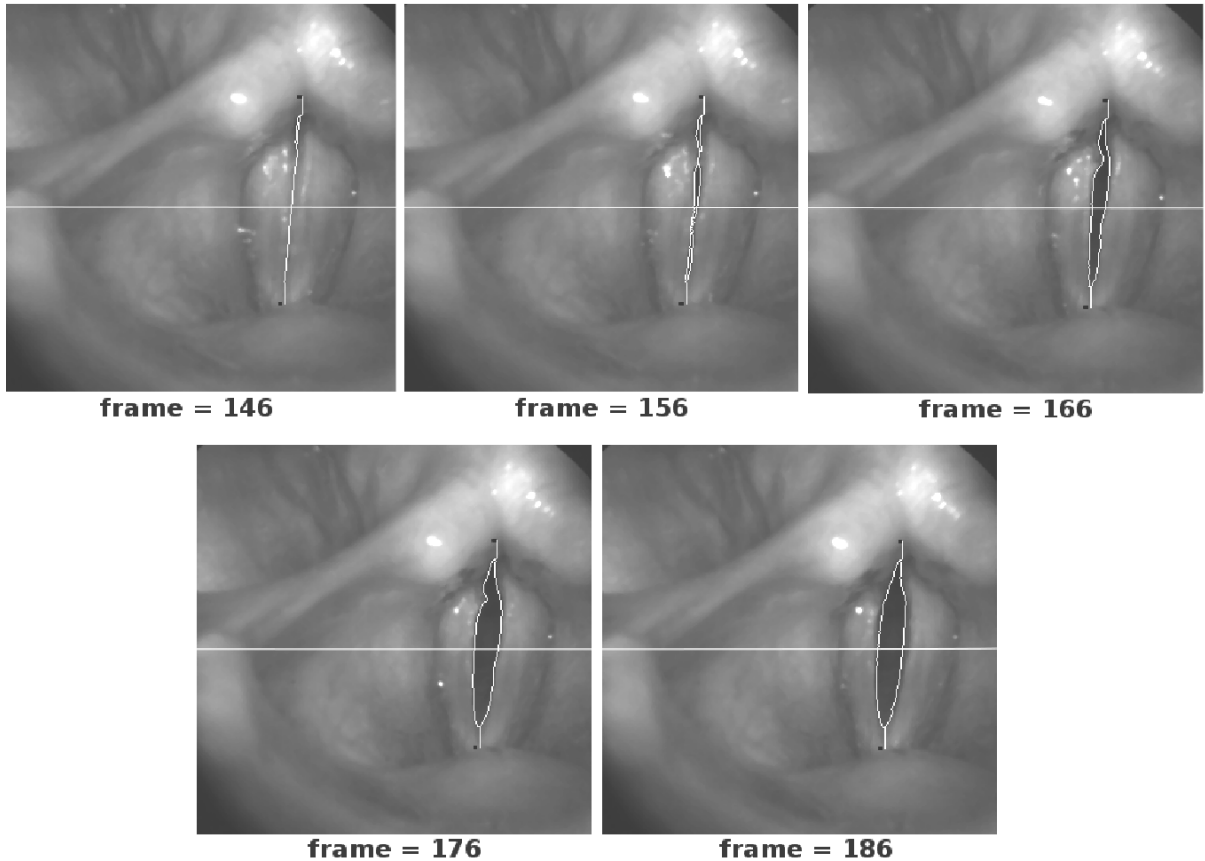}
\caption{A subject's left and right vocal folds delineated using the temporal-snake transform, with white contours imposed on the vocal fold boundaries. This particular sequence shows the transition from a closed to an opened glottis. The horizontal white line across the vocal folds has been superimposed to demonstrate the approximate middle kymographic scan line between the top and bottom of the vocal folds. The brightness in these images has been increased for printing.}
\label{fig:Fig3.2}
\end{figure}

\subsection{A Snake in the Turing Machine's Shadow (Contribution 1)}

In the previous subsection, we introduced snakes as a computerized spline deformed in the spatial domain of a digital image in order to minimize energy and result in feature extraction. Our arguments in this section are closely related to the snake model. It is our belief that such arguments may hold for many other computational models in image analysis and merits continued investigation. Before a snake can be deformed to minimize its energy, it must have a proper initialization. Although in practice snake initialization can be tricky and often times not well defined, we ignore the need of initialization in the proof of the following theorem in order to simplify the proof construction, since it can easily be shown that there are an exponential number of legal configurations the snaxels of a snake can be in before applying the snake-deformation algorithm. This does not detract from the general argument, rather it makes it more accessible to the reader. In practice, snake-initialization is typically achieved via information granted from a higher level process, operator specification, or some other computable function as defined by the application.

Recall, traditionally, a snake will have at least three parameters, $\alpha$ a weight on the spline's continuity, $\beta$ a weight on the spline's rigidity or difficulty in bending, and $\gamma$ a weight governing the strength of or how reliable the data term is. The fact of the matter is, given parameters, a snake's energy can be minimized, however, the problem of snake-parameter selection is a computationally intractable problem. This is not to say that an approximation, or some heuristic or empirical human made observations on the optimal parameters or optimal parameter range can be devised for a particular data set or even data type, rather, there is no way to define the optimal parameters via an algorithm, even if an ideal initialization is always available.

\newtheorem{theorem3.1}[theorem]{Theorem}

\begin{theorem3.1}
\noindent Proof: \textnormal{The language of snake parameters is undecidable.} \\

\textnormal{Recall, the language of snake parameters is given by the following set}

\[
P = \{\alpha, \beta, \gamma \hspace{4pt} | \hspace{4pt} \alpha, \beta, \gamma \in \mathbb{Q}\},
\]

\noindent \textnormal{that is, the language of triplets of rational fractions. \hspace{4pt} $\therefore$}

\[
\forall \hspace{4pt} p \in P, \hspace{4pt} \exists \hspace{4pt} x, y, z \in Q, \hspace{4pt} \textnormal{such that} \hspace{4pt} p = \{x, y, x\}
\]

\[
\Rightarrow p \in \{\mathbb{Q} \times \mathbb{Q} \times \mathbb{Q}\}
\]

\[
\Rightarrow P = \{\mathbb{Q} \times \mathbb{Q} \times \mathbb{Q}\}
\]

\noindent \textnormal{via Lemma $2$.$1$, a bijection can be used to show $\left| \mathbb{Q} \right| = \left| \mathbb{N} \right|$,}

\[
\Rightarrow P = \{\mathbb{N} \times \mathbb{N} \times \mathbb{N}\}
\]

\noindent \textnormal{Now, we must construct a bijection $g:\mathbb{N} \times \mathbb{N} \times \mathbb{N} \rightarrow \mathbb{N}$. We could construct another multidimensional table and enumerate all entries of $\mathbb{N \times N \times N}$ along with all the entries of $\mathbb{N}$ to pair the elements, however, we can take a more direct approach by taking advantage of function composition to derive a closed form formula. Using the bijection $f$ established in lemma $2$.$2$, the formula for the bijection $g(a, b, c) = d$ is as follows:}

\begin{align*}
g(a, b, c) &= f(a, f(b, c))  \\
	&= \frac{(a + f(b, c))(a + f(b, c) + 1)}{2} + f(b, c),
\end{align*}

\noindent \textnormal{thus, a bijection $g:\mathbb{N \times N \times N}\rightarrow \mathbb{N}$ has been established, which in turn illustrates $\left|P\right| = \left|\mathbb{N}\right|$. Now, via lemma $2$.$3$, a language and its complement must both be Turing-recognizable in order for the language to be decidable. To summarize, in the case of the language of snake parameters, the language itself can be reduced to the natural numbers, however, the complement of this set, the real numbers, is not even Turing recognizable!} \\

\noindent $\therefore$ \hspace{4pt} \textnormal{the language of snake parameters} $P$ \textnormal{is undecidable}. \qed

\end{theorem3.1}

Again, it is quite possible to determine some heuristic or employ empirical observations on the parameters or set of parameters that lead a snake to ideal convergence on a sequence of images, however, it is impossible to write a computable function to calculate optimal snake parameters, no such function exists. 

\newtheorem{corollary3.1}[theorem]{Corollary}

\begin{corollary3.1}
\textnormal{by Theorem $3$.$1$, automated snake algorithms that require computing updated snake parameters are intractable by means of deterministic computation.}
\end{corollary3.1}

\noindent In practice, when dealing with image sequences, there arise occasions in which the same parameters cannot be used for all snakes we wish to deform within the series. Especially when images exhibit sharp changes in brightness or contrast, or cases in which a snake is in regions of strong noise and a lack of noise simultaneously. We note that the dynamic programming algorithm for snake deformation does allow adjusting $\alpha$, $\beta$, and $\gamma$ on a per-snaxel basis, however, the problem of selecting parameters on a per-snaxel basis is still undecidable.

\section{Discussion}

In the previous section, we proved theorem $3$.$1$, the language of snake parameters is not Turing-decidable, given a particular snake instance, suitable parameters cannot be determined algorithmically. However, this result should not come as a big surprise since such models in image analysis employ variational frameworks for regularizing a solution, to make ill-posed problems more manageable. In constraint optimization problems, \textit{hard-constraints} can significantly reduce the cardinality of the set of possible solutions, since \textit{all} solutions will have to conform to a particular structure. Whereas, \textit{soft-constraints} are constraints only on the cost function employed for the optimization process. In regards to snake models, the parameters $\alpha$ and $\beta$ for specifying a contour's internal energy are regularization terms, in fact, they are soft-constraints, they do not convey any additional information that \textit{all} solutions must obey, rather information regarding the solution for a specific instance of the problem. The parameters have been introduced to make the energy optimization solvable in the first place. 

Thus, the difficulty in solving the ill-posed energy minimization problem has been casted from the problem of even being able to find a solution to a particular problem (by introducing regularization parameters), to the problem of finding suitable parameters to be used in finding a solution to a particular problem instance, which in turn implies, herein lies the problem! In general, it is the non-uniqueness of the minimal energy contour produced by algorithms such as the snake model that make these problems difficult to automate, especially in the context of sharp variations or changes when traversing image sequences.

\subsection{A Human Being is a Deciding Being}

Although fully automated biomedical interpretation is desired, it is a long term goal. Furthermore, semiautomated methods may in fact prove to be more useful until a general mathematical theory explaining human perception and understanding of imagery has been developed, or the problem of simulating human intelligence via computer has been solved. A skilled professional such as a clinician or surgeon could take advantage of semiautomated methods that can potentially automate a vast majority of the monotonous and redundant tasks in image interpretation, rather than the automation acting as a second opinion or a replacement. With semiautomated image analysis techniques, the user of a software tool is able to adjust the process until they obtain a desirable result. One of the co-creators of the original snake model has gone through great effort to summarize the different types of snake and deformable models for medical image analysis, concluding semiautomatic and interactive methods will remain dominant in practices for years to come \cite{mcinerney:96}. 

We wish to take the argument a step further. As semiautomated methods for biomedical image analysis dominate the field, these methods will be the \textit{key} to designing fully automated methods, and may even serve as a gateway to developing simulations of human-like intelligence via computer. In the case of the temporal-snake transform \cite{moukalled:09}, after a few attempts a user would converge to the set of parameters that yield the best result for a particular image sequence. This should not come as a surprise as humans are undoubtedly great deciders of computationally hard problems. 

Let us digress for a moment and take note that \textit{supervised learning} and \textit{semi-supervised learning} algorithms, are methods which take advantage of data labeled by a human that is representative of the phenomenon of interest, by employing variational techniques for performing classification or optimizing parameters that yield desirable results \cite{james:03}. Such training algorithms not only require human interaction to produce samples of what is desired by the analysis, but tuning the parameters again can be quite time consuming, and also require large amounts of time for off-line training. Furthermore, such methods may require retraining upon the introduction of new features and or data that deviates greatly from the anticipated input, and suffer when the data contains large amounts of redundant information. 

In recent years, \textit{unsupervised} and \textit{deep learning} algorithms have gained substantial amounts attention, due to the desire of determining relationships in big data and to the widespread use of the internet and search engines. Deep learning methods are often casted as off-line unsupervised machine learning problems which seeks to determine the global optimum in pattern classification without human interaction, by taking advantage of the statistical power of extremely large data sets to determine a relationship among samples \cite{bengio:09}. Deep learning methods are promising, yet again, the off-line training can be quite time consuming and require many computational resources, and even the most state-of-the art deep learning techniques achieve recognition rates of $74\%$ - $82\%$ \cite{le:2012}. In the context of medical image interpretation, such resource requirements and performance cannot be tolerated. Due to the requirement of enormous amounts of data to take advantage of the statistical power of such methods, and due to the lack of accessibility of several large databases of medical images of the same type of anatomical structure, application of these approaches is infeasible. Without large networks of data readily available between medical image collaborators and machine learning experts, employing these learning techniques is difficult for the time being.

Returning back to the underlying theme of this section of the text, aside from the medical image interpretation problem, another great example of computationally hard problems are \textit{video games}, which are in fact deterministic computer simulations. Despite their novelty and being ridiculed as entertainment for children, understanding such interactive systems may play a major role in many aspects of developing automated human intelligence. After some finite number of attempts, a human player of a video game will converge to the solution of a particular puzzle, dungeon, level, or even complete the game. Yet, in recent years, researchers have demonstrated the algorithmic complexity of automatically playing such games. Automating a player that obeys the rules of simulated worlds of several video games and models of game play that have been around since the late $1980$'s is computationally intractable \cite{demaine:2003, aloupis:2012, viglietta:2012}. 

Within the blink of an eye, a player must adjust their strategies and continue to play all while acquiring and processing several stimuli simultaneously. Similarly, while using semiautomated medical imaging tools, the operator must adapt, and proceed in a fashion that yields desirable results. By taking advantage of the human computer, the brain, and limiting the amount of interaction necessary, it is our belief that a human being can accurately and quickly extract features of interest from medical images. And upon proper feature extraction, any additional analysis can be fully automated.

\subsection{Automata are All Around Us and Also Within Us (Contribution 2)}

Indeed, automata are all around us and some within us. For example, anyone who has visited a supermarket is immediately greeted by an automaton upon entering. The little proximity sensor that resides above the entrance that detects when a person is standing near the doorway is a finite automaton that automatically controls the opening and closing of the sliding doors \cite{sipser:06}. Although finite automata are simple reflex agents, when a system comprised of several automata interacting with one another is in motion, the system exhibits complex and even chaotic behavior. \textit{Cellular automata} first introduced by Ulman, von Neumann, and Zuse in the $1950$'s are discrete grids with a finite automaton existing at each grid point, where each automaton's state is visible to itself and its neighbors \cite{wolfram:84}. 

In some cases, cellular automata can be viewed as discretization of partial differential equations in time. For example, \textit{lattice-gas cellular automata} and \textit{lattice Boltzmann} are cellular automata in applications of fluid and gas flows that have been used to solve and derive macroscopic properties of the Navier-Stokes equations and the discrete Boltzmann equation, respectively \cite{wolf-gladrow:2000}. In addition to the behavior of fluids and gases, cellular automata appear in a wide array of places in nature. Cellular automata can be used to describe plant growth, plant branching patterns, growth of antlers in animals, shell geometry, shell visual pattern generation, various genetic programs (e.g. DNA replication), simple animal behaviors, animal pigmentation patterns (e.g. butterflies, fish, snakes, tigers, and zebras), and even tissue growth (e.g. bones and tumors) \cite{wolfram:2002}. 

Wolfram, one of the first authors to widely adopt and build upon cellular automata argues much like how the telescope helped shape astronomy, and how the microscope helped shape biology, computer technology will help shape engineering and mathematics \cite{wolfram:2002}. We agree with this view, and would like to take it a step further, more specifically, automation and computation will indeed help shape biomedical image analysis, especially when the tool is used properly. As engineers and scientists, we must know the limitations of computational tools we brandish and develop intuition for recognizing when our tools can be naturally extended to solve additional problems.

Following in the footsteps of Wolfram, our long term goal is to determine cellular automata rules that accurately generate vocal fold vibrations. We take the first steps towards this goal by defining a non-context-free language that synthesizes magnitude of image gradients for normal vocal fold vibrations as would be derived from kymographic image data. Let $\exists$ a finite alphabet $\Sigma = \{\bigcirc, \square, \bigtriangleup\}$, where 

\begin{align*}
\bigcirc &= \{\textnormal{vocal folds opening / opened}\} \\
\square &= \{\textnormal{vocal folds closing}\} \\
\bigtriangleup &= \{\textnormal{vocal folds in contact / closed}\}
\end{align*}

\noindent Now, loosely speaking, a non-context-free language that can be decided by a Turing machine in quadratic time that describes normal vocal-fold vibrations with respect to the information contained within temporally rich kymographic image sequences would be

\begin{equation}
V_{\textnormal{NCFL}} = (\bigcirc^j \square^k \bigtriangleup^i)^* \cap (\square^k \bigtriangleup^i \bigcirc^j)^* \cap (\bigtriangleup^i \bigcirc^j \square^k)^*,
\end{equation}

\noindent where $i, j, k > 1$, $c_1 i \approx c_2 j \approx c_3 k$, and $c_1, c_2, c_3 > 0$. As seen in Figs. $5$-$9$, using a subset of strings generated from the last term of Eq. ($5$), and a pseudo-random number generator, a kymographic images of healthy vibration for short vocal folds as taken at the middle kymographic scan line are synthesized.

\begin{figure}[!ht]
\centering
\includegraphics[width=14cm]{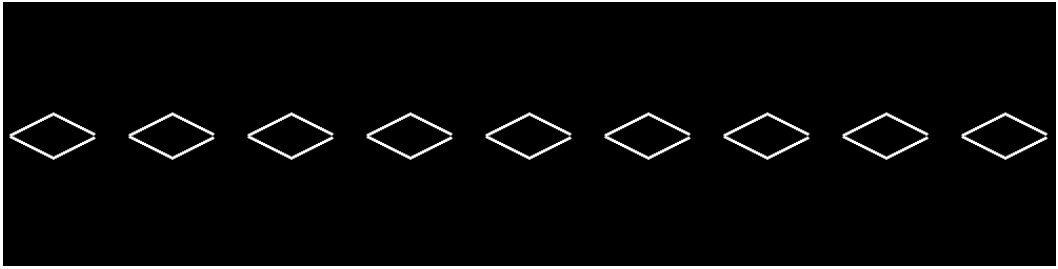}
\caption{Computer synthesized kymographic image gradient magnitude for short vocal folds extracted at the middle scan line. This synthesized kymogram resembles a habitual phonation.}
\label{fig:Fig4.1}
\end{figure}

\begin{figure}[!ht]
\centering
\includegraphics[width=14cm]{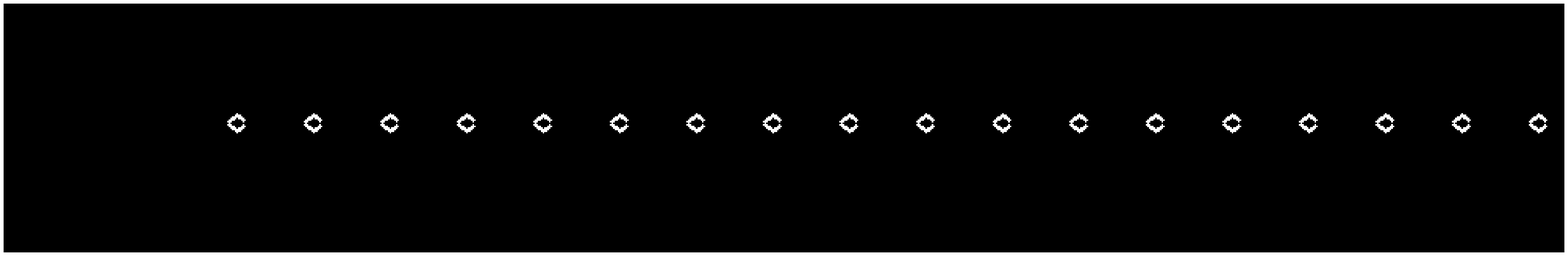}
\caption{Computer synthesized kymographic image gradient magnitude for short vocal folds extracted at the middle scan line. This synthesized kymogram resembles a high phonation.}
\label{fig:Fig4.2}
\end{figure}

\begin{figure}[!ht]
\centering
\includegraphics[width=14cm]{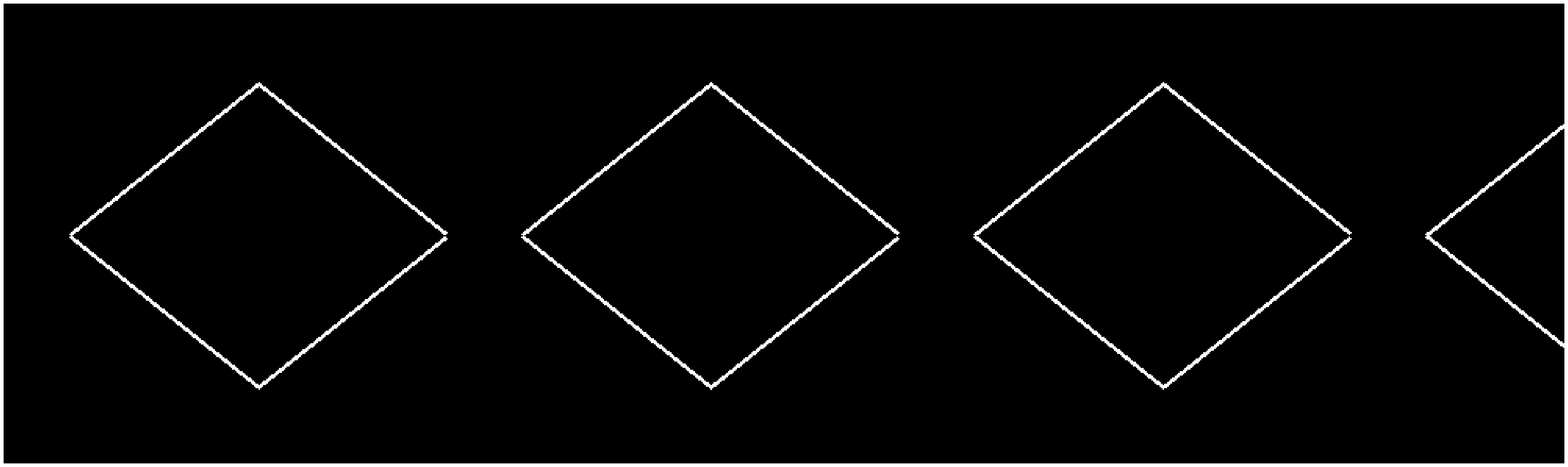}
\caption{Computer synthesized kymographic image gradient magnitude for short vocal folds extracted at the middle scan line. This synthesized kymogram resembles a breathy phonation.}
\label{fig:Fig4.3}
\end{figure}

\newpage

\begin{figure}[!ht]
\centering
\includegraphics[width=14cm]{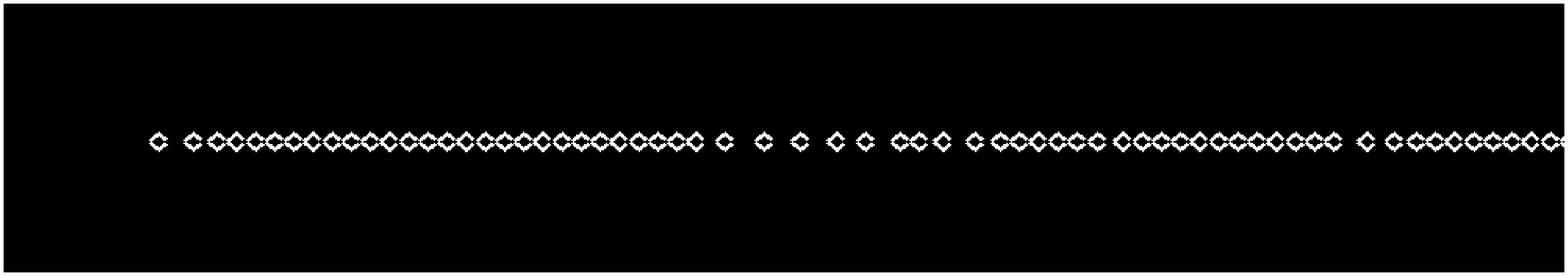}
\caption{Computer synthesized kymographic image gradient magnitude for short vocal folds extracted at the middle scan line. This synthesized kymogram  resembles a falsetto phonation.}
\label{fig:Fig4.4}
\end{figure}

\begin{figure}[!ht]
\centering
\includegraphics[width=14cm]{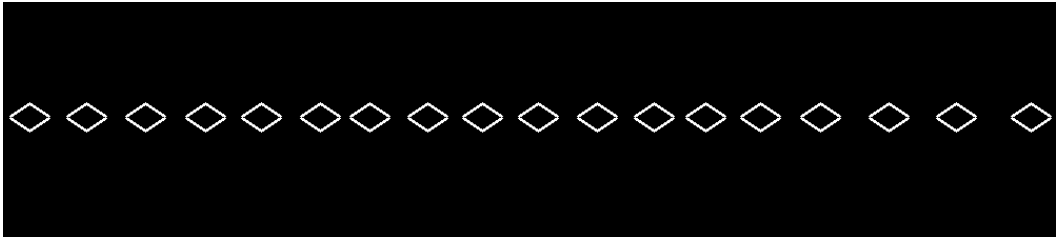}
\caption{Computer synthesized kymographic image gradient magnitude for short vocal folds extracted at the middle scan line. This synthesized kymogram resembles a habitual phonation.}
\label{fig:Fig4.5}
\end{figure}


\newtheorem{conjecture4.1}[theorem]{Conjecture}
\begin{conjecture4.1}
$\exists$ \textnormal{a $2$D cellular automata that evolves in time capable of simulation normal vocal fold vibrations.}
\end{conjecture4.1}

\noindent Some classes of cellular automata are \textit{Turing-complete}, which makes them capable of simulating Turing machine programs. Furthermore, some cellular automata are reversible. This has deep implications, given cellular automata rules that are representative of biomedical phenomena. It may be possible to construct analysis tools that incorporate these rules and automata to make more difficult problems not only manageable, but provide a unique solution for quantifying the phenomenon of interest.


\section{Conclusion}
In this work, we take the first steps towards developing automata rules for generating languages that explain physiological characteristics of vocal fold vibrations with respect to laryngeal imaging and its applications. It is our strong belief that not only do such approaches require continued investigation, but can be extended to other areas of physiological signal processing. Furthermore, we should be able to take advantage of individual finite automata that model a deterministic subset of the actions space of anatomical structures to later define cellular automata that evolve in time to study natural medical phenomena. 

We live in a universe were objects change according to patterns and rules, we call these rules the laws of nature \cite{sagan:89}. Identifying such rules in the context of biomedical applications will allow us to establish computable languages capable of describing healthy behaviors of anatomical structures, and possibly better automated classification of pathological conditions. Before we can make the leap, we must harness satisfactory semiautomated analysis tools that have been decomposed in such a way that reduces human errors, and where human interaction is minimal and has been isolated to a very specific part of the framework. These achievements may yield better understanding of human perception, and to algorithms to simulate human intelligence.

\section*{Acknowledgment}
A special thanks to Dr. Dimitar Deliyski and the NIH R$01$ grant DC$007640$ on the efficacy of laryngeal imaging. Thanks to the University of South Carolina Department of Computer Science. This work was supported by the unfaltering love and encouragement of family and friends throughout the process.

\bibliographystyle{abbrv}
\bibliography{cited}

\end{document}